\documentclass{article}

\usepackage[preprint]{neurips_2025}

\usepackage{textcomp}
\usepackage[utf8]{inputenc} 
\usepackage[T1]{fontenc}    
\usepackage{hyperref}       
\usepackage{url}            
\usepackage{booktabs}       
\usepackage{amsfonts}       
\usepackage{nicefrac}       
\usepackage{microtype}      
\usepackage{xcolor}         

\usepackage{multirow}
\usepackage{multicol}
\usepackage{makecell}
\usepackage{color}
\usepackage{pifont}
\usepackage{amsfonts}

\usepackage{bm}
\usepackage{colortbl}
\usepackage{graphicx}
\usepackage{amsmath}
\usepackage{amsthm}
\usepackage{amssymb}
\usepackage{booktabs}
\usepackage{algorithm}
\usepackage{algorithmic}
\usepackage{subcaption}
\usepackage{mathrsfs}
\usepackage{wrapfig}
\usepackage{xcolor}
\usepackage{enumitem}
\usepackage{setspace}

\DeclareMathOperator{\bilinear}{BILINEAR}

\newcommand{\name}{MSF}
\title{\name: Efficient Diffusion Model via Multi-Scale Latent Factorization}

\author{
Haohang Xu\textsuperscript{1} \footnotemark[1] \; 
Longyu Chen\textsuperscript{1,2} \footnotemark[1] \;
Yichen Zhang\textsuperscript{4} \footnotemark[1] \;
Shuangrui Ding\textsuperscript{3} \;
Zhipeng Zhang\textsuperscript{4} \footnotemark[2] \\
\textsuperscript{1} Huawei Inc. \; \textsuperscript{2} Huazhong University of Science \& Technology \; \\
\textsuperscript{3} The Chinese University of Hong Kong \; \\
\textsuperscript{4} School of Artificial Intelligence, Shanghai Jiao Tong University \\
}

\begin{document}

\maketitle
\begin{abstract}

    While diffusion-based generative models have made significant strides in visual content creation, conventional approaches face computational challenges, especially for high-resolution images, as they denoise the entire image from noisy inputs. This contrasts with signal processing techniques, such as Fourier and wavelet analyses, which often employ hierarchical decompositions. Inspired by such principles, particularly the idea of signal separation, we introduce a diffusion framework leveraging multi-scale latent factorization. Our framework uniquely decomposes the denoising target, typically latent features from a pretrained Variational Autoencoder, into a low-frequency base signal capturing core structural information and a high-frequency residual signal that contributes finer, high-frequency details like textures. This decomposition into base and residual components directly informs our two-stage image generation process, which first produces the low-resolution base, followed by the generation of the high-resolution residual. Our proposed architecture facilitates reduced sampling steps during the residual learning stage, owing to the inherent ease of modeling residual information, which confers advantages over conventional full-resolution generation techniques. This specific approach of decomposing the signal into a base and a residual, conceptually akin to how wavelet analysis can separate different frequency bands, yields a more streamlined and intuitive design distinct from generic hierarchical models. Our method, \name\ (Multi-Scale Factorization), demonstrates its effectiveness by achieving FID scores of 2.08 ($256\times256$) and 2.47 ($512\times512$) on class-conditional ImageNet benchmarks, outperforming the DiT baseline (2.27 and 3.04 respectively) while also delivering a $4\times$ speed-up with the same number of sampling steps. Thus, our \name\ effectively balances generation quality with computational efficiency through its innovative base-residual decomposition strategy. 
    
  
\end{abstract}

{
    \renewcommand{\thefootnote}{\fnsymbol{footnote}}
    \footnotetext[1]{Equal contributions.}
    \footnotetext[2]{Corresponding author}
}

\section{Introduction}
Diffusion models have recently demonstrated impressive performance across a range of generative tasks, including image~\cite{saharia2022photorealistic,sdxl,liu2024playground,dai2023emu}, video~\cite{ho2022imagen, zhang2024show, singermake, blattmann2023align, blattmann2023stable}, and audio synthesis~\cite{audio-diffwave, Grad-TTS, Guided-TTS, audio-WaveGrad2I}. These models iteratively transform random Gaussian noise into high-quality outputs by progressively applying denoising steps. Despite their effectiveness, conventional diffusion models operate directly on high-resolution images throughout the denoising process, which results in significant computational demands due to the multi-step reverse sampling. This issue is especially pronounced in architectures with $O(n^2)$ computational complexity, such as DiT~\cite{dit} and FiT~\cite{fit, fitv2}.


To accelerate diffusion models, significant efforts have recently been devoted. One line of research~\cite{ode-dpm, zheng2023dpm, zhou2024fast, kim2024distilling, lu2022dpm} focuses on expediting the reverse sampling process by employing deterministic differential equation solvers. These methods reformulate the reverse diffusion process as either a Stochastic Differential Equation (SDE)~\cite{sde} or an Ordinary Differential Equation (ODE)~\cite{liuflow}, thereby enabling the use of various high-order solvers to speed up sampling. However, while these solver-based sampling methods accelerate the generation process, they typically do not address the computational overhead inherent in the diffusion model's architecture itself. Recently, another alternative branch, namely step distillation techniques~\cite{salimans2022progressive, Nguyen_2024_CVPR, lin2024sdxllightning, yan2024perflow}, has also emerged to expedite sampling. Nevertheless, the distillation process inherently introduces substantial additional training costs and significantly increases GPU memory demands, as it necessitates the concurrent utilization of both student and teacher models during training. These considerations naturally lead to the question that ``\textit{Is it possible to innovate the architecture of diffusion models to simultaneously enhance both speed and performance without incurring significant additional costs?}''

\begin{figure}
    \centering
    \includegraphics[width =\textwidth]{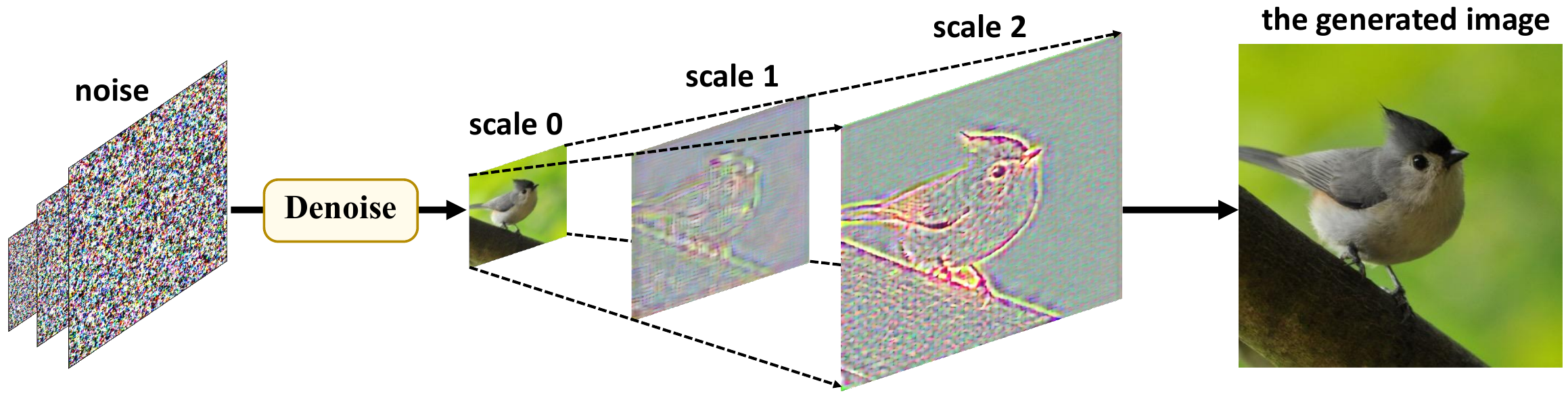}
    \caption{An illustration of \name, where diffusion models generate hierarchical representations at multiple scales. The multi-scale representations are progressively accumulated to produce the final output. At each scale, the denoising target (scale 0\textasciitilde 2) is derived from latents extracted by an off-the-shelf, pretrained Variational Autoencoder (VAE).}
    \label{fig: intro}
    \vspace{-10pt}
\end{figure}

Our answer is affirmative. Rather than concentrating solely on sampling strategies, we address the acceleration challenge from a novel perspective by decomposing the overall image generation task into several sub-tasks, each of which can be addressed with a simpler architecture. Specifically, we decompose the generation process into multiple scales, with each scale representing a distinct level of information. The initial scale, which has the lowest resolution, contains the most fundamental information, similar to the low-frequency component in wavelet analysis. Compared to full-resolution generation, these low-resolution components can be trained or sampled more efficiently. Subsequent scales capture increasingly finer residual information between the full-resolution input and the lower-resolution components at each scale, as shown in Fig.~\ref{fig: intro}. As a common wisdom, residual information is often easier to model, allowing us to use  fewer sampling steps to process these high-resolution residual components. Based on this, we propose a multi-scale training and sampling pipeline, employing a single DiT backbones to generate images.

To sum up, we decompose the complex full-resolution generation task into two simplified stages: (i) generating a low-resolution base component, and (ii) generating multiple high-resolution residual components. Through experiments, \name\ achieves an excellent balance between the generation quality and computational overhead, showing that the combined cost of these two stages is significantly lower than that of the original full-resolution generation task under the premise of better generation quality. In particular, \name\ achieves an FID of 2.47 on ImageNet 512 $\times$ 512, outperforming the baseline DiT~\cite{dit} with an FID of 3.04, while delivering a speedup of over 4$\times$ for the same sampling steps. Moreover, the proposed approach is orthogonal to, and potentially complementary with, existing techniques for accelerating diffusion models. For instance, \name\ can effectively integrate with step distillation, further boosting the sampling speed, as detailed in Sec.~\ref{sec:ex}.

The key contributions of this paper are as follows:
\begin{itemize}[leftmargin=.2in]
  \item We propose a hierarchical target decomposition strategy for diffusion models, which systematically factorizes the input into a foundational low-resolution base component and multiple high-resolution residual components, each capturing distinct levels of image detail.
  
\item We design a corresponding multi-scale training and sampling pipeline, wherein a unified Diffusion Transformer (DiT) backbone is employed to generate these decomposed target components at their respective spatial scales.

  \item Extensive experiments validate the effectiveness of \name, demonstrating notable improvements in computational efficiency, \textit{i.e.,} delivering a speedup of over $4\times$ in sampling than the baseline DiT, while achieving superior generation quality, especially at higher resolutions such as $512 \times 512$.
\end{itemize}

\section{Related Work}
\paragraph{Diffusion-based Image Generation.}  

Diffusion-based models~\cite{ho2020denoising, song2020denoising, song2020score, nichol2021improved} emerge as a prominent class of generative methods. These models function by learning to reverse a predefined forward noising process, progressively transforming a noisy input into a clean sample through iterative denoising steps. While this iterative refinement mechanism enables the generation of high-fidelity images and offers superior training stability compared to Generative Adversarial Networks (GANs)~\cite{gan, stylegan, kang2023scaling, mirza2014conditional, chen2016infogan}, it also introduces considerable computational overhead during the sampling phase. To address this challenge, Latent Diffusion Models (LDMs)~\cite{rombach2022high} are introduced, which operate by performing the diffusion process within the compressed latent space of a Variational Autoencoder (VAE~\cite{vae}). The significant spatial compression provided by the VAE encoder, typically an $8\times8$ downsampling, leads to substantial improvements in both training and sampling efficiency. This enhanced computational efficiency is instrumental in scaling diffusion models to larger and more complex architectures. Following the LDM paradigm, a subsequent body of work~\cite{liu2024playground, dai2023emu, sdxl, chen2023pixart, chen2024pixart1, chen2024pixart2} focuses on further advancing these models to produce images of even higher quality and resolution, thereby expanding the frontiers of diffusion-based visual synthesis.


\paragraph{Diffusion Backbone.} Diffusion-based generative models predominantly utilize two primary architectural frameworks. Initial methodologies, exemplified by works such as~\cite{ho2020denoising, rombach2022high}, employ traditional U-Net structures as their backbone. To enhance the scalability of these models, DiT~\cite{dit} introduces the use of Transformers, specifically a Vision Transformer, as the backbone for Latent Diffusion Models (LDMs), replacing the conventional U-Net. Subsequent research further explores Transformer based architectures. For instance, MDT~\cite{gao2023masked} proposes a masked latent modeling technique to improve the learning of internal contextual relationships within images by Diffusion Probabilistic Models (DPMs). U-ViT~\cite{bao2023all} adopts a unified token-based representation for all inputs and incorporates long skip connections. Building upon the DiT framework, SiT~\cite{sit} introduces an interpolation scheme and investigates various rectified flow configurations. FiT~\cite{fit} explores the generation of images at arbitrary resolutions and aspect ratios, and its successor, FiTv2~\cite{fitv2}, incorporates additional engineering optimizations to further improve convergence speed and performance. Nevertheless, these aforementioned approaches generate full resolution images at a single scale, which incurs significant computational overhead. In this work, drawing inspiration from frequency decomposition principles and building upon the established DiT architecture, we propose \name, an efficient diffusion model achieved through multi-scale latent factorization.

\paragraph{Multi-stage Diffusion Models.} Cascaded diffusion models, also termed multi-stage diffusion models, are introduced and widely adopted for generating high-quality visual content~\cite{ho2022cascaded, saharia2022photorealistic, ho2022imagen, zhang2024show, singermake}. This process typically commences with a base diffusion model producing low-resolution images and subsequently cascades to generate high-resolution visual content through a series of supplementary diffusion models and interpolation networks~\cite{blattmann2023align, blattmann2023stable}. However, in these frameworks, each stage undertakes a complete generation task, thereby incurring considerable computational costs. PDD~\cite{liu2023pyramid} applies pyramidal discrete diffusion models to synthesize large-scale 3D scenes under resource constraints, utilizing a coarse-to-fine approach for progressive upscaling and employing scene subdivision techniques. DoD~\cite{yue2024diffusion} proposes a multi-stage generation framework that enhances diffusion models with visual priors, wherein the initial stage performs standard class-conditional generation, and subsequent stages extract visual priors by encoding the outputs of the preceding stage to guide the diffusion model. Nevertheless, DoD's multi-stage diffusion model maintains a uniform resolution across all stages, and its initial stage encounters challenges in generating high-resolution images from coarse and limited class information. In contrast, our approach partitions the generation process into multiple scales and progressively refines residual details between the full-resolution input and its lower-resolution counterparts at each stage. This multi-scale residual information is more tractable to model compared to direct full-resolution generation, allowing for a simpler architecture that improves computational efficiency for both training and sampling.

\section{Method}
In this section, we present a comprehensive description of the proposed \name\ method. First, we provide an overarching overview of \name\ to contextualize its components in Sec.~\ref{subsec: overview}. Subsequently, we detail the key modules of \name\ as follows: (i) the extraction of multi-scale latent representations using a pretrained variational autoencoder (VAE) in Sec.~\ref{subsec: multi-scale VAE}; (ii) the multi-scale diffusion network that generates latent representations at each scale in Sec.~\ref{subsec: diff-loss}; and (iii) the sampling process, which aggregates the residual latent representations from all levels to produce the final output in Sec.~\ref{subsec: accumulate sampling}.

\subsection{An Overview of \name}
\label{subsec: overview}
Given an input image $\boldsymbol{x} \in \mathbb{R}^{H \times W \times 3}$, a multi-scale VAE encodes it into a set of tokens $\left\{ \boldsymbol{f_0}, \boldsymbol{f_1}, \ldots, \boldsymbol{f_N} \right\}$, where each token $\boldsymbol{f_i} \in \mathbb{R}^{h_i\times w_i \times C}$ represents a distinct scale. After we obtain the multi-scale latents $\left\{\boldsymbol{f_0}, \boldsymbol{f_1}, \ldots, \boldsymbol{f_N} \right\}$, a straightforward approach is using all preceding scales to predict the ground truth latent of the current scale: 
\begin{align} \label{eq: naive-ar} 
    p(\boldsymbol{f_0}, \boldsymbol{f_1}, \ldots, \boldsymbol{f_N}) = \prod_{i=0}^{N}p(\boldsymbol{f_i}|\boldsymbol{f_0}, \ldots, \boldsymbol{f_{i-1}}) 
\end{align}
where $p(\cdot)$ indicates a diffusion-based next-scale prediction network, conditioned on the preceding scale inputs. In this paper, each $\boldsymbol{f_i}$ essentially represents a decomposition of the visual signal. To model Eq.~\ref{eq: naive-ar} more effectively, we scale all latents from previous scales to a unified scale and aggregate them as a condition for predicting the latent at the current scale, or as a prior.
We will delve into this process for the priors in Sec.~\ref{subsec: diff-loss}.

\subsection{Multi-scale Variational Autoencoder} 
\label{subsec: multi-scale VAE}
Given a pretrained VAE encoder $\mathcal{E}$, we can derive the latent representation $\boldsymbol{\hat{f}} \in \mathbb{R}^{h \times w \times C}$ via $\boldsymbol{\hat{f}} = \mathcal{E}(\boldsymbol{x})$. Various methods are available to factorize the single-scale latent into multi-scale latents:

\paragraph{Scaling Latent} A straightforward method for obtaining multi-scale latents involves scaling the latent representation via down-sampling interpolation. This operation can be directly applied to either the VAE latents $\boldsymbol{\hat{f}}$ or raw RGB inputs $\boldsymbol{x}$. We formulate these two methods as follows, respectively:
\begin{align}
    &\boldsymbol{f_i} = \bilinear\ (\boldsymbol{\hat{f}}, h_i, w_i) \\
    &\boldsymbol{f_i} = \mathcal{E}(\bilinear\ (\boldsymbol{x}, H_i, W_i))
\end{align}
where $h_i(H_i)$ and $w_i(W_i)$ denote the target height and width.

\begin{figure}
    \centering
    \includegraphics[width=\textwidth]
    {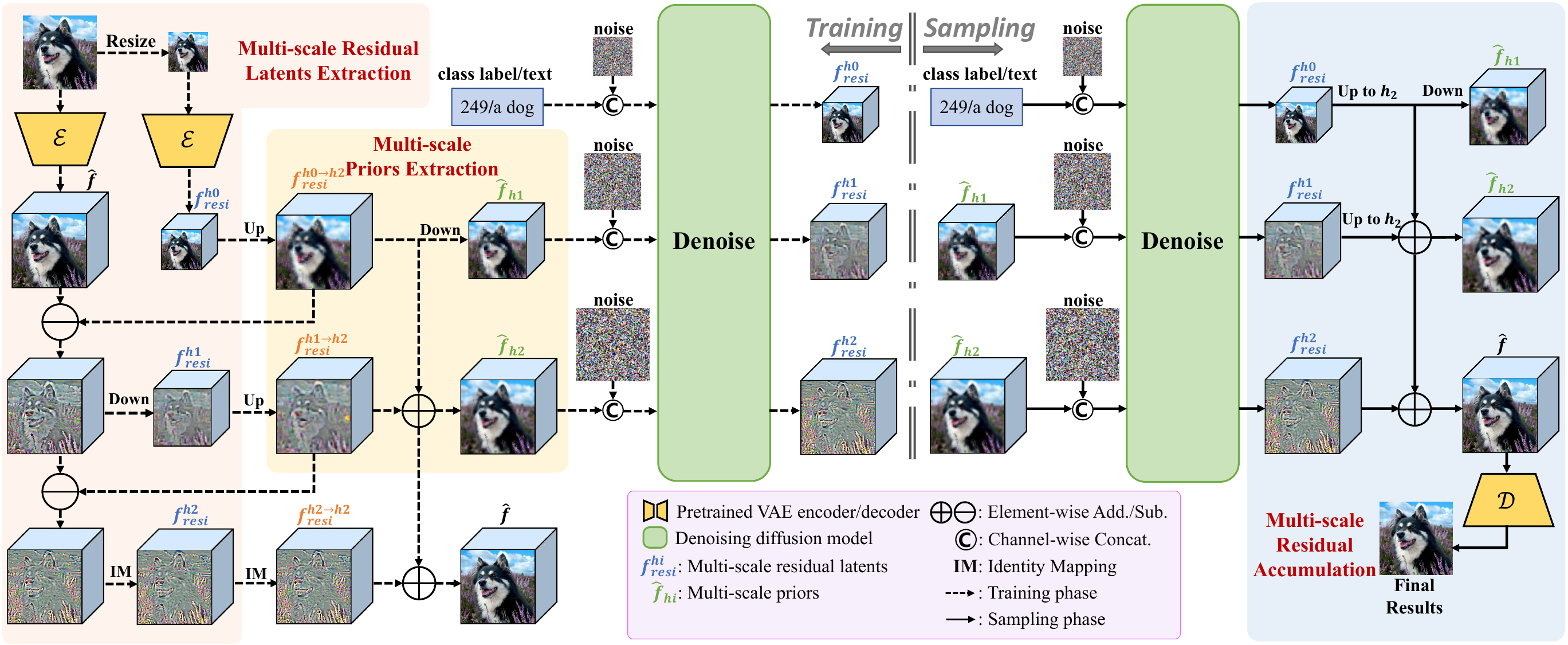}
    \caption{An overview of the proposed \name. \textbf{Left:} The figure on the left illustrates the process of extracting multi-scale residual latents $\boldsymbol{f_{resi}^{h_i}}$ using the pretrained VAE. \textbf{Middle:} The middle depicts the construction process of multi-scale priors $ \boldsymbol{{\hat f}_{h_i}}$ and how the diffusion model takes $ \boldsymbol{{\hat f}_{h_i}}$ as input to generate the ground-truth residual latents $ \boldsymbol{f_{resi}^{h_i}}$. \textbf{Right:} During sampling, we first generate the residual latents $ \boldsymbol{f_{resi}^{h_i}}$, and then perform residual accumulation to obtain the $\boldsymbol{{\hat f}_{h_i}}$, which serves as input for the next-scale prediction.}
    \label{fig: arch}
    \vspace{-10pt}
\end{figure}

\begin{wrapfigure}{r}{0.6\textwidth}
    \vspace{-23pt}
    \begin{minipage}{0.6\textwidth}
      \begin{algorithm}[H]
        \caption{Multi-scale Residual Latents Extraction}
        \label{alg:multi_scale_residual_latents_extraction}
        \small
        \begin{algorithmic}[1] 
          \STATE \textbf{Inputs: } high and low resolution image pair $(\boldsymbol{I_h}, \boldsymbol{I_l})$, resolutions $(h_i,w_i)_{i=0}^{N}$;
          \STATE \textbf{Encode: } $\boldsymbol{\hat{f}} = \mathcal{E}(\boldsymbol{I_h})$, $\boldsymbol{f_0} = \boldsymbol{f_{resi}^{h_0}} = \mathcal{E}(\boldsymbol{I_l})$;
          \STATE \textbf{Initialize: } $\boldsymbol{R} = \{\boldsymbol{f_{resi}^{h_0}}\}$;
          \FOR{$i=0$ to $N-1$} 
            \STATE $\boldsymbol{\hat{f}} = \boldsymbol{\hat{f}} - \text{Up}(\boldsymbol{f_{resi}^{h_i}}, h_N, w_N)$;
            \STATE $\boldsymbol{f_{resi}^{h_{i+1}}} = \text{Down}(\boldsymbol{\hat{f}}, h_{i+1}, w_{i+1})$;
            \STATE $\boldsymbol{R} = \boldsymbol{R} \cup \{ \boldsymbol{f_{resi}^{h_{i+1}}}\}$;
          \ENDFOR
          \STATE \textbf{Return: } multi-scale residual latents $\boldsymbol{R}$;
        \end{algorithmic}
      \end{algorithm}
   \end{minipage}
   \vspace{-10pt}
\end{wrapfigure}
\paragraph{Scaling Residual Latent.} The aforementioned scaling methods generate multi-scale latents from the original single-scale latent through direct interpolation. Though these approaches allow each latent at different scales to represent information from images of varying resolutions, they do not establish an explicit relationship between the generated multi-scale latents. As a result, the multi-scale latents may independently capture different types of information, without any coherent interaction.

To ensure the hierarchical relationship of the extracted multi-scale latents, we
propose ``scaling of residual latents'', which leverages the concept of residual learning, a fundamental technique in the field of deep learning~\cite{he2016deep}. Within our VAE framework employing residual learning, the first step involves extracting the base component $\boldsymbol{f_0}$ from the latent representation $\boldsymbol{\hat{f}}$. The base component should capture the most fundamental, low-level information of $\boldsymbol{\hat{f}}$, akin to the low-frequency components in Fourier decomposition. We evaluate two methods for obtaining the base component: (1) directly downsampling $\boldsymbol{\hat{f}}$ in the latent space, and (2) first downsampling the input image and then generating its VAE latent. We find that the first method, downsampling in the latent space, results in significant information loss. In contrast, the second method, downsampling the input image before VAE encoding, preserves most of the information contained in $\boldsymbol{\hat{f}}$. Consequently, \name\ adopts the second~method.

\paragraph{Multi-scale Residual Latents Extraction.} Given the original latent representation $\boldsymbol{\hat{f}}$ and its primary component $\boldsymbol{f_0}$ (denoted as $\boldsymbol{f_{resi}^{h_0}}$ in Fig.~\ref{fig: arch}), we first compute the residual latent at the initial scale as $\boldsymbol{\hat{f}} - \text{Up}(\boldsymbol{f_0})$, where $\text{Up}(\cdot)$ represents the upsampling operation employed to match the spatial dimensions of $\boldsymbol{\hat{f}}$ and $\boldsymbol{f_0}$. To further decompose the first-scale residual latent $\boldsymbol{\hat{f}} - \text{Up}(\boldsymbol{f_0})$, we derive its fundamental component by applying downsampling $\boldsymbol{f_{resi}^{h_1}} = \text{Down}(\boldsymbol{\hat{f}} - \text{Up}(\boldsymbol{f_0}))$, where $\text{Down}(\cdot)$ represents the downsampling operation. The subsequent steps follow a similar procedure. A detailed description of this process is provided in Alg.~\ref{alg:multi_scale_residual_latents_extraction}. During training, the extracted multi-scale residual latents $\boldsymbol{R} = \{\boldsymbol{f_{resi}^{h_i}}\}_{i=0}^{N}$ serve as the ground truth for each scale prediction.

\begin{wrapfigure}{r}{0.6\textwidth}
\vspace{-23pt}
\begin{minipage}{0.6\textwidth}
\begin{algorithm}[H]
\caption{Multi-scale Priors Extraction}
\label{alg:multi_scale_input_priors_extraction}
\small
\begin{algorithmic}[1] 
\STATE \textbf{Inputs:} multi-scale residual latents $\boldsymbol{R} = \{\boldsymbol{f_{resi}^{h_i}}\}_{i=0}^{N}$;
\STATE \textbf{Initialize: } $\boldsymbol{\hat{f}} = 0$, $\boldsymbol{P} = \{\}$;
\FOR{$i=0$ to $N-1$} 
    \STATE $\boldsymbol{f_{resi}^{h_i \rightarrow h_N}} = \text{Up}(\boldsymbol{f_{resi}^{h_i}}, h_N, w_N)$;
    \STATE $\boldsymbol{\hat{f}} = \boldsymbol{\hat{f}} + \boldsymbol{f_{resi}^{h_i \rightarrow h_N}}$;
    \STATE $\boldsymbol{\hat{f}_{h_{i+1}}} = \text{Down}(\boldsymbol{\hat{f}}, h_{i+1}, w_{i+1})$;
    \STATE $\boldsymbol{P} = \boldsymbol{P} \cup \{ \boldsymbol{\hat{f}_{h_{i+1}}}\}$;
\ENDFOR
\STATE \textbf{Return: } multi-scale priors $\boldsymbol{P}$;
\end{algorithmic}
\end{algorithm}
\end{minipage}
\vspace{-13pt}
\end{wrapfigure}

\subsection{Multi-scale Diffusion Network}
\label{subsec: diff-loss}
\paragraph{Multi-scale Priors Extraction.} 
In our framework, the prior $\boldsymbol{\hat{f}_{h_i}}$ represents the cumulative latent representations from scales 0 to $i-1$.
Notably, the initial scale has no prior and directly produces a low-resolution primary signal from class labels $c$. It is natural for the model to execute sequentially in the sampling phase. Whereas during the training phase, we utilize teacher-forcing~\cite{goodman-etal-2020-teaforn}, a strategy that trains a model by using ground truth data as input at each step instead of its own predictions, to guide the model in parallel training. The process of extracting multi-scale priors reflects the inverse of the decomposition performed during multi-scale residual latents extraction on the original latent representation $\boldsymbol{\hat{f}}$. Beginning at scale 0, $\boldsymbol{\hat{f}}$ accumulates residuals $\boldsymbol{f_{resi}^{h_i}}$ progressively across scales. It is important to note that these residuals must be upsampled to $\boldsymbol{f_{resi}^{h_i \rightarrow h_N}}$ to match the spatial dimensions of $\boldsymbol{\hat{f}}$. The prior $\boldsymbol{\hat{f}_{h_i}}$ is obtained by downsampling $\boldsymbol{\hat{f}}$, which has integrated information from scales 0 through $i-1$, to match the dimensions of the current scale. As illustrated in Fig.~\ref{fig: arch}, the aggregation of latent representations across all scales enables the reconstruction of $\boldsymbol{\hat{f}}$. The detailed procedure for obtaining the multi-scale priors $\boldsymbol{P} = \{\boldsymbol{\hat{f}_{h_i}}\}_{i=1}^{N}$ during the training phase is elaborated in Alg.~\ref{alg:multi_scale_input_priors_extraction}.

\paragraph{Residual-based Next-scale Prediction.} The priors, obtained via Alg.~\ref{alg:multi_scale_input_priors_extraction}, serve as the condition to generate the residual target at the current scale. Specifically, we use the rectified flow~\cite{liuflow} model due to its stability during training and fast sampling speed. 
Rectified flow aims to learn a velocity field ${v}$, which transforms random noise $\boldsymbol{z_1} \sim \mathcal{N}(0,I)$ into the ground truth distribution $\boldsymbol{z_0} = \boldsymbol{{f}_{resi}^{h_i}} \sim \pi_{data}$:
\begin{align}
    \mathcal{L} = \sum_{i=0}^{N}\int_0^1 \mathbb E[|| (\boldsymbol{z_1} -\boldsymbol{f_{resi}^{h_i}}) - v_{\theta}(\boldsymbol{z_t},t | c, \boldsymbol{f_{resi}^{h_{i-1}}})||^2] dt
\end{align}
where $N$ is the number of total scales, $c$ is the class label, and $\boldsymbol{z_t} = t \cdot \boldsymbol{z_1} + (1-t) \cdot \boldsymbol{f_{resi}^{h_i}}$. The latents at each scale are encoded with a segment embedding $e_i$ in the transformer to differentiate information across scales. After training, an ODE function is used to sample $\boldsymbol{f_{resi}^{h_i}}$ at different scales:
\begin{align}
    {\rm d} {\boldsymbol{z_t}} &= { v}_{\theta}(\boldsymbol{z_t}, t | c, \boldsymbol{f_{resi}^{h_{i-1}}})\rm{d}t
\end{align}

\begin{wrapfigure}{r}{0.5\textwidth}
\vspace{-23pt}
\begin{minipage}{0.5\textwidth}
\begin{algorithm}[H]
\caption{Autoregressive sampling}
\label{alg:autoregressive_sampling}
\small
\begin{algorithmic}[1] 
\STATE \textbf{Inputs: } class label $c$, resolutions $(h_i, w_i)_{i=0}^{N}$;
\STATE \textbf{Initialize: } $\boldsymbol{\hat{f}} = 0$;
\FOR{$i=0$ to $N$} 
    \IF{$i == 0$}
        \STATE $\boldsymbol{f_{resi}^{h_i}} \sim {v}_{\theta}(\boldsymbol{{z}_t}, t | c)$;
    \ELSE
        \STATE $\boldsymbol{f_{resi}^{h_i}} \sim {v}_{\theta}(\boldsymbol{z_t}, t | c, \boldsymbol{\hat{f}_{h_i}})$;
    \ENDIF
    \STATE $\boldsymbol{\hat{f}} = \boldsymbol{\hat{f}} + \text{Up}(\boldsymbol{f_{resi}^{h_i}}, h_N, w_N)$;
    \IF{$i < N$}
        \STATE $\boldsymbol{\hat{f}_{h_{i+1}}} = \text{Down}(\boldsymbol{\hat{f}}, h_{i+1}, w_{i+1})$;
    \ENDIF
\ENDFOR
\STATE \textbf{Decode: } $\boldsymbol{\hat{I_h}} = \mathcal{D}(\boldsymbol{\hat{f}})$;
\STATE \textbf{Return: } generated high-resolution image $\boldsymbol{\hat{I_h}}$;
\end{algorithmic}
\end{algorithm}
\end{minipage}
\vspace{-15pt}
\end{wrapfigure}

\subsection{Multi-scale Accumulate Sampling} 
\label{subsec: accumulate sampling}
The sampling process begins with the class label $c$ and computes the residual latent representation $\boldsymbol{f_{resi}^{h_i}}$ for each scale $i$ in a manner similar to autoregressive methods. The residual latents are accumulated to obtain the final latent representation $\boldsymbol{\hat{f}}$. Specifically, the initial scale generates a low-resolution primary signal $\boldsymbol{f_{resi}^{h_0}}$ based on $c$, while subsequent scales produce high-resolution residual signals conditioned on the class label $c$ and prior $\boldsymbol{\hat{f}_{h_i}}$. Each computed residual signals $\boldsymbol{f_{resi}^{h_i}}$ is upsampled to $(h_N, w_N)$ and accumulated into $\boldsymbol{\hat{f}}$. And then $\boldsymbol{\hat{f}}$, containing the cumulative results of all current scales, is downsampled to $(h_{i+1}, w_{i+1})$ and serves as the prior for the next scale. After incorporating the residual latent of the last scale, the VAE decoder generates the image from $\boldsymbol{\hat{f}}$. The complete procedure for sampling is detailed in Alg.~\ref{alg:autoregressive_sampling}.

\section{Experiments}
\label{sec:ex}
In this section, we conduct experiments on ImageNet~\cite{deng2009imagenet} at the resolutions of $256\times256$ and $512\times512$. We first evaluate our method with both the Fréchet Inception Distance (FID) and Inception Score (IS) using standard evaluation protocols. Subsequently, we compare the sampling time of \name\ with the baseline method DiT, demonstrating the effectiveness of \name. We also analyze the performance of \name\ with several detailed ablation studies in Sec.~\ref{sec:A & D}. 


\subsection{Experimental Setup}
We train diffusion models using the ImageNet 2012 training dataset. Following Stable Diffusion~\cite{rombach2022high}, we employ a VAE encoder with a downsampling ratio of 8 to extract latent representations of input images. We set the number of scales as 2 for both $256\times256$ and $512\times512$ generation: the first scale is fixed at $192\times192$ and the second is determined by the target resolution, either $256\times256$ or $512\times512$. 
We provide two variants of \name\ with different model sizes, \name-B (Base) with 19 DiT-transformer blocks and \name-L (Large) with 28 DiT-transformer blocks. The entire training process consists of two stages. In stage 0, we first train scale 0 independently until convergence. Once converged, we proceed to stage 1, where scale 1 is trained jointly with scale 0. During this stage, the training losses from both scales are simply summed and backpropagated together. The iteration counts reported in Tab.~\ref{tab:main} refer to the total number of iterations across both stages.

Due to GPU memory limitations, we set the batch size to 1024 for scale 0 and 512 for scale 1. The learning rate is fixed at $1 \times 10^{-4}$. As shown in Tab.~\ref{tab:main}, the number of training iterations required for scale 1 is approximately $1/4$ of that for scale 0, since scale 1 only models the residual information.




\begin{table}[t]
\renewcommand\arraystretch{0.9}
\centering
\small
\caption{Quantitative Comparison of Class-Conditional Generative Methods on ImageNet $256\times256$ and $512\times512$. In this comparison, ``$\downarrow$'' indicates that lower values are preferable, while ``$\uparrow$'' indicates that higher values are better. We report results on Fréchet Inception Distance (FID), Inception Score (IS), Precision (Pre), and Recall (Rec). Note that $^*$ represents \name\ being trained with a patch size of 4 at scale 1, which enables faster training and sampling compared with the default patch size of 2.}
\label{tab:main}
\scriptsize
\vspace{5pt}
\resizebox{1\linewidth}{!}{
\begin{tabular}{lccc ccc}
\toprule
{Model}   & {Iters (K)} & {Para}  & {FID $\downarrow$} & {IS $\uparrow$} & {Pre $\uparrow$} & {Rec $\uparrow$} \\
\bottomrule \\[-4pt]
\textbf{ImageNet $256\times256$} & \\[-2.7pt]
\midrule
VQVAE-2~\cite{Razavi2019GeneratingDH}  &2207  & 13.5B  & 31.11  & $\sim$45 & 0.36  & 0.57  \\
VDM++ ~\cite{kingma2024understanding}  &$-$ & 2B & 2.12 &267.7 &$-$ &$-$  \\
ViTVQ~\cite{Yu2021VectorquantizedIM}  & 950 & 1.7B   & 4.17  & 175.1  & $-$  & $-$   \\
VQGAN~\cite{esser2021taming}   & $-$ & 1.4B   & 15.78 & 74.3   & $-$  & $-$   \\

ADM~\cite{dhariwal2021diffusion}    & 1980 & 554M   & 10.94 & 100.98   & 0.69 & 0.63 \\
ADM-G~\cite{dhariwal2021diffusion}  & 1980 & 554M   &4.59  & 186.7 &0.82 &0.52  \\
LDM-8-G~\cite{rombach2022high}   & 4800  & 506M  & 7.76   & 209.52  & 0.84 & 0.35 \\

LDM-4-G~\cite{rombach2022high}  & 178  & 400M & 3.60  & 247.67  & 0.87  & 0.48  \\
GIVT-Causal ~\cite{tschannen2025givt}  & $-$ & 304M  &5.67  & $-$  &0.75  &0.59 \\
MaskGIT~\cite{Chang2022MaskGITMG}   & 1500 & 227M   & 6.18  & 182.1  & 0.80 & 0.51  \\
CDM~\cite{ho2022cascaded}     & 1600 & $-$   & 4.88  & 158.7   & $-$  & $-$  \\ 
\arrayrulecolor{gray}\cmidrule(lr){1-7}
DiT-L~\cite{dit}    & $-$ & 458M   & 5.02  & 167.2  & 0.75 & 0.57  \\

\cellcolor{lightgray!40}{\name-B} 
&\cellcolor{lightgray!40}{1100 + 125}
&\cellcolor{lightgray!40}{368M}   
&\cellcolor{lightgray!40}{2.76}  
&\cellcolor{lightgray!40}{218.3}      
&\cellcolor{lightgray!40}{0.84}  
&\cellcolor{lightgray!40}{0.55}    \\

DiT-XL~\cite{dit}   & 7000 & 675M  & 2.27  & 278.2  & 0.83 & 0.57  \\

\cellcolor{lightgray!40}{\name-L} 
&\cellcolor{lightgray!40}{1025 + 200}
&\cellcolor{lightgray!40}{680M}   
&\cellcolor{lightgray!40}{2.08}  
&\cellcolor{lightgray!40}{245.5}      
&\cellcolor{lightgray!40}{0.83} 
&\cellcolor{lightgray!40}{0.57}   \\

\bottomrule \\[-4pt]
\textbf{ImageNet $512\times512$} & \\[-2.7pt]
\midrule
VDM++ ~\cite{kingma2024understanding}  & $-$ & 2B & 2.65 &278.1 &$-$ &$-$ \\

ADM~\cite{dhariwal2021diffusion}    & 1940  & 554M  &23.24 &58.06 &0.73 &0.60  \\
ADM-G~\cite{dhariwal2021diffusion}  & 1940  &554M  &7.72  &172.71  &0.87  &0.42  \\

\arrayrulecolor{gray}\cmidrule(lr){1-7}

DiT-XL~\cite{dit}  & 3000 & 675M  & 3.04 & 240.82 & 0.84 & 0.54  \\

\cellcolor{lightgray!40}{\name-L$^*$} 
&\cellcolor{lightgray!40}{1025 + 450} 
&\cellcolor{lightgray!40}{680M}   
&\cellcolor{lightgray!40}{2.58}  
&\cellcolor{lightgray!40}{252.8}      
&\cellcolor{lightgray!40}{0.83}  
&\cellcolor{lightgray!40}{0.54}   \\
\cellcolor{lightgray!40}{\name-L} 
&\cellcolor{lightgray!40}{1025 + 200} 
&\cellcolor{lightgray!40}{680M}   
&\cellcolor{lightgray!40}{2.47}
&\cellcolor{lightgray!40}{249.1}      
&\cellcolor{lightgray!40}{0.83}  
&\cellcolor{lightgray!40}{0.53}   \\

\bottomrule
\end{tabular}
}
\vspace{-5pt}
\end{table}


\subsection{Comparison with State-of-the-arts.} 

\begin{figure}[t]
    \centering
    \includegraphics[width=\textwidth]{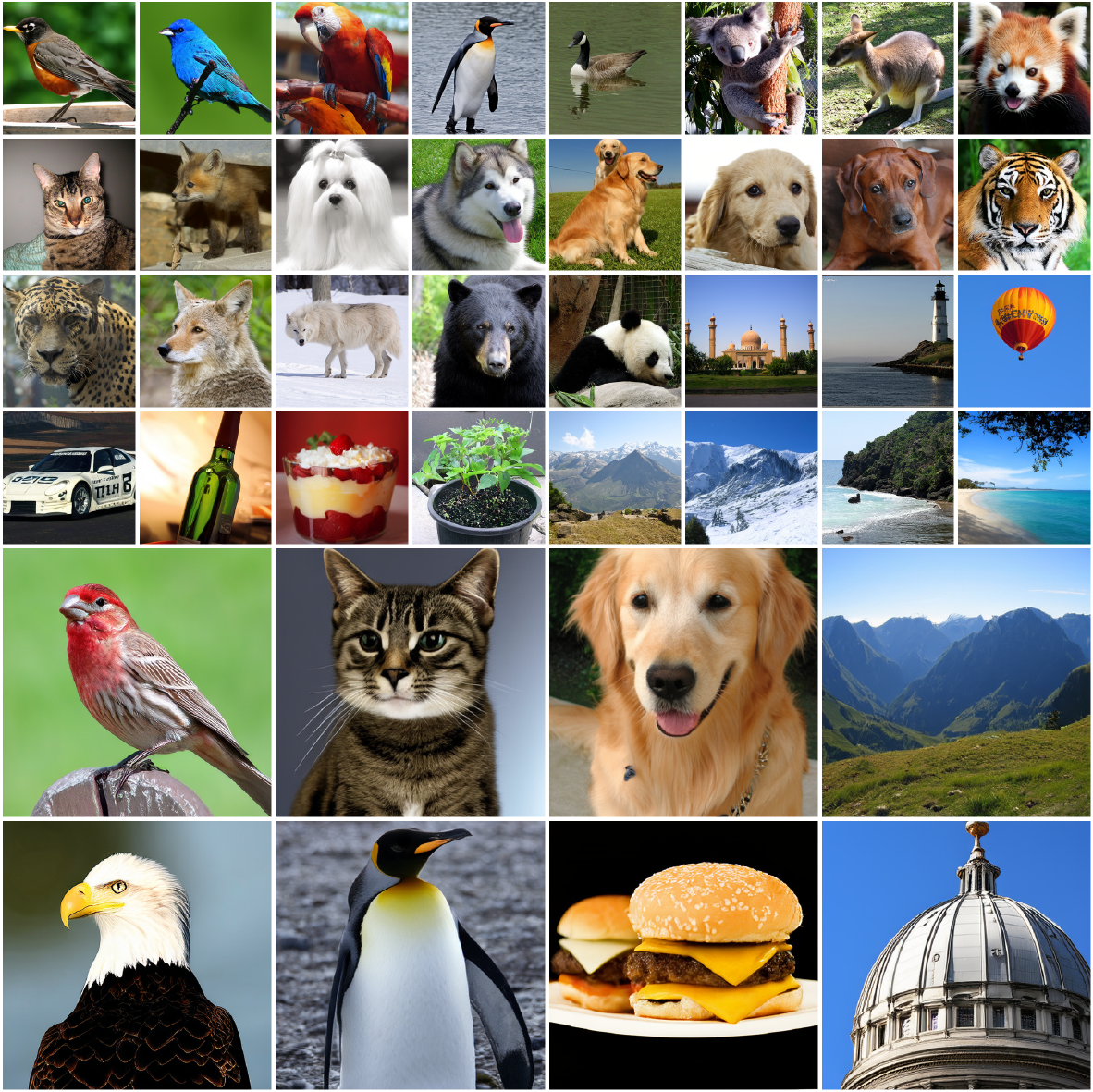}
    \caption{Some generated $256\times256$ and $512\times512$ samples by \name\ trained on ImageNet.}
    \label{fig: gen}
\vspace{-5pt}
\end{figure}

\textbf{Results on ImageNet.} This section presents a comparison of our method against other works on the ImageNet dataset, evaluated at $256\times256$ and $512 \times 512$ resolutions.
We evaluate \name\ using several standard metrics, including Fréchet Inception Distance (FID), Inception Score (IS), Precision, and Recall. For a fair comparison, we adopt evaluation scripts from ADM~\cite{dhariwal2021diffusion} and report results on 50K generated samples across 1000 ImageNet categories. We set the sampling steps for the two scales to 100 and 20, and the classifier-free guidance scales to 1.3 and 1.0 (\textit{i.e.}, without CFG). As shown in Tab.~\ref{tab:main}, \name\ achieves an excellent balance between the generation quality and computational overhead, with FID scores of 2.08 and 2.47 on ImageNet $256 \times 256$ and $512 \times 512$, respectively, clearly outperforming DiT-L and DiT-XL baselines. We also train the \name-L$^*$ variant for $512\times512$ generation, which uses a transformer patch size of 4 at scale 1. In this setting, the number of default image tokens at scale 1 is reduced from $(512 / 8 / 2)^2 = 1024$ to $(512 / 8 / 4)^2 = 256$ (where 8 is the VAE downsample size). Remarkably, even with the token count reduced to $1/4$, \name\ maintains strong generation performance, which significantly improves both training and sampling speed.
 We also provide the visualizations of samples generated by \name\ in Fig.~\ref{fig: gen}.

\paragraph{Sampling Speed Compared with DiT} 
Beyond superior generation quality, \name\ also accelerates sampling significantly. For high-resolution ($512 \times 512$) image synthesis, we first perform an efficient sampling stage at scale 0 ($192 \times 192$ resolution), followed by only a few steps to generate the residual information at scale 1. Notably, we observe that classifier-free guidance is not necessary at scale 1, eliminating the need to double the batch size for conditional and unconditional generations.

\begin{wraptable}{r}{0.55\textwidth}
\vspace{-1pt}
\setlength{\tabcolsep}{3pt}
\small
\centering
\caption{The generation results with few sampling steps of \name-L on ImageNet $512\times512$. The sampling steps of \name\ is measured as $s_0$ + $s_1$, where $s_0$ is the steps of scale 0 and $s_1$ is the steps of scale 1. The sampling time is measured on a A800 GPU with batch size of 64.}
\label{tab:few-step gen}
\vspace{-2pt}
\begin{tabular}{l l ccc}
\toprule
Model   & Steps & FID-5K & FID-50K & Time(s)\\
\midrule
DiT-XL &250 &$-$ &3.04 &816 \\
DiT-XL &100 &$-$ &$-$  &326 \\
\midrule
\name-L   & 100 + 20  & 5.07  & 2.47  &76 \\
\name-L   & 100 + 10  & 5.23  &2.58  & 70 \\
\name-L   & 100 + 4   & 6.54  & 3.19 & 68 \\
\midrule
\name-L   & 50  + 20  &5.32     &2.51 & 48  \\
\name-L   & 50  + 4   &6.67     &3.24 & 36 \\
\arrayrulecolor{gray}\cmidrule(lr){1-5}
\name-L$^*$   & 100 + 20   &5.33   & 2.58 & 67 \\
\midrule
\textbf{Step Distillation} & \\[-2.5pt]
\midrule
\name-L     & 10 + 10    & 6.79  &3.91 & 11 \\
\name-L      & 5 + 10   & 7.37  &4.35  & 8 \\
\name-L      & 5 + 4   & 8.16  &4.90  & 4\\
\bottomrule
\end{tabular}
\vspace{-15pt}
\end{wraptable}

As shown in the top section of Tab.~\ref{tab:few-step gen}, \name-L generates 64 images at $512 \times 512$ resolution in \texttt{76} seconds, whereas DiT requires \texttt{326} seconds with similar sampling steps, showing our MSF is more than $\textbf{4}\times$ faster than DiT. Even increasing the sampling steps of DiT to 250, our model still demonstrates better performance and runs at $\textbf{11}\times$ faster. Notably, the sampling steps at scale 1 can be significantly reduced. For example, using only 4 steps at scale 1 still achieves FID comparable to DiT.


As mentioned earlier, in the original \name\ model, the majority of sampling cost lies in the scale 0 generation. 
As previously discussed, our approach is complementary to other acceleration techniques. To further improve efficiency, we integrate piecewise rectified flow (PeRFlow)~\citep{yan2024perflow} as a plug-and-play accelerator for scale 0 sampling (see ''step distillation'' in the Tab.~\ref{tab:few-step gen}). Following PeRFlow’s methodology, we divide the scale 0 sampling trajectory into 4 time windows and apply a reflow operation to straighten the flow trajectories within each window. After applying step distillation at scale 0, \name-L achieves a sampling speed of \texttt{16}~$imgs/sec$ while maintaining competitive generation performance. Visualizations of the distilled \name\ outputs are provided in the supplementary material.



\section{Analysis and Discussion}
\label{sec:A & D}

\begin{wraptable}{r}{0.55\textwidth}
\vspace{-12pt}
\renewcommand\arraystretch{1.05}
\setlength{\tabcolsep}{3.5pt}
\centering
\small
\caption{
{FID-5K vs. CFG scale.} (a) CFG at scale 1 is fixed to 1.0 while varying at scale 0. (b) CFG at scale 0 is fixed to 1.3 while varying at scale 1.
    }
\label{tab:diff-cfg-step}
\begin{subtable}{.22\textwidth}
\vspace{-2pt}
\center
\caption{}
\label{tab:diff-cfg-step:a}
\begin{tabular}{ccc}
    \toprule
    \multicolumn{2}{c}{CFG} &\multirow{2}*{FID-5K} \\
    \cmidrule(lr){1-2} scale 0 &scale 1  \\
    \midrule
    1.0 &1.0 & 9.77 \\
    1.1 &1.0 & 6.88 \\
    1.2 &1.0 & 5.51 \\
    1.3 &1.0 & 5.23 \\
    1.4 &1.0 & 5.48 \\
    1.5 &1.0 & 6.16 \\
    \bottomrule
\end{tabular}
\end{subtable}\hfill
\begin{subtable}{.28\textwidth}
\vspace{-1pt}
\center
\caption{}
\label{tab:diff-cfg-step:b}
\begin{tabular}{ccc}
    \toprule
    \multicolumn{2}{c}{CFG} &\multirow{2}*{FID-5K} \\
    \cmidrule(lr){1-2} scale 0 &scale 1  \\
    \midrule
    1.3 &1.0 &5.23 \\
    1.3 &1.1 &5.23 \\
    1.3 &1.2 &5.23\\
    1.3 &1.3 &5.22\\
    1.3 &1.4 &5.22\\
    1.3 &1.5 &5.22\\
    \bottomrule
\end{tabular}
\end{subtable}
\vspace{-5pt}
\end{wraptable}

\paragraph{Effects of Classifier-Free Guidance Scale.} 
Classifier-free guidance~\cite{ho2022classifier} (CFG) is a core technique to enhance the generative quality of diffusion models by interpolating between conditional and unconditional outputs. We evaluate the outputs of each scale under varying CFG individually, as shown in Tab.~\ref{tab:diff-cfg-step:a}. With the CFG scale of scale 1 fixed at 1.0, we evaluate FID-5K for scale 0 under different CFG, finding the best result at 1.3 and the performance of scale 0 generation primarily impacted by CFG. Then, fixing the CFG scale of scale 0 at 1.3, we execute the same operation for scale 1, observing that scale 1 achieves satisfactory results even without CFG (Tab.~\ref{tab:diff-cfg-step:b}). Since CFG requires doubling the forward passes of the inputs and introduces additional sampling costs, we omit CFG for scale 1 generation (\textit{i.e.}, CFG=1.0 for scale 1) to achieve a 2$\times$ speedup in sampling.


\paragraph{Effects of the Nmuber of Scales.}
In \name, we propose dividing the image generation into multiple stages, each addressing a specific scale. To explore how the number of scales affects the model's generation capability, we conduct experiments with three scales (192 $\to$ 384 $\to$ 512) besides the two-scale experiments. Tab.~\ref{tab:diff-num_scales} shows CFG of the first scale 0, with other scales set to CFG $= 1$, and the sampling steps of all scales. The results indicate that the generation quality degrades compared to the two-scale setting. We hypothesize that this is due to error accumulation as the number of scales increases. Therefore, we recommend using a two-scale configuration for \name.

\begin{wraptable}{r}{0.58\textwidth}
\vspace{-13pt}
\setlength{\tabcolsep}{3.5pt}
\small
\centering
\caption{Performance comparison across different numbers of scales, The patch size of scale 1 and scale 2 are set as 4, the same as the \name-L$^*$ setting in Tab.~\ref{tab:main}}
\label{tab:diff-num_scales}
\vspace{-2pt}
\begin{tabular}{lcccc}
\toprule
Model  & CFG  &Steps & Scales  & FID-5K \\
\midrule 
\name-L$^*$    & 1.3  & 100 + 10   & 192 $\to$ 512   & 5.44 \\
\arrayrulecolor{gray}\cmidrule(lr){1-5}
\name-L$^*$  &  1.2  &100 + 10 + 10    & 192 $\to$ 384 $\to$ 512   & 9.44  \\
\name-L$^*$  &  1.3  &100 + 10 + 10    & 192 $\to$ 384 $\to$ 512   &  8.29 \\
\name-L$^*$  &  1.4  &100 + 10 + 10    & 192 $\to$ 384 $\to$ 512   &  7.81 \\
\name-L$^*$  &  1.5  &100 + 10 + 10    & 192 $\to$ 384 $\to$ 512   & 7.84  \\
\name-L$^*$  &  1.6  &100 + 10 + 10    & 192 $\to$ 384 $\to$ 512   & 8.09  \\
\bottomrule
\end{tabular}
\vspace{-10pt}
\end{wraptable}

\vspace{10pt}
\section{Conclusion}
\label{sec:conclu}
 In this paper, we propose a versatile and efficient image generation framework that decomposes the full-resolution image generation process into a multi-scale and residual-based, progressive hierarchy. Each scale within the hierarchy captures distinct image details, with the framework learning both a fundamental low-resolution component and several residual high-resolution components. This structure enables to model the multi-scale information progressively, leading to substantial efficiency gains without compromising output quality. Through extensive experiments, we demonstrated that such multi-scale factorization approach enhances the generative quality, achieving higher fidelity with fewer computational resources than traditional methods, validating the effectiveness and efficiency of \name\ in high-resolution image~synthesis. Currently, \name\ is exclusively evaluated on class-conditional image generation tasks. In future work, we plan to extend its framework to text-to-image generation, enabling multimodal conditional synthesis.


\clearpage
\bibliographystyle{plain}
\bibliography{myref}

\end{document}